\documentclass{article}

 \PassOptionsToPackage{numbers, compress}{natbib}
     \usepackage[preprint]{neurips_2019}

\usepackage{cite}
\usepackage{amsmath,amssymb,amsfonts}
\usepackage{algorithm}
\usepackage{algorithmic}

\usepackage{graphicx}
\usepackage{textcomp}
\usepackage{xcolor}
\usepackage{enumitem}
\def\BibTeX{{\rm B\kern-.05em{\sc i\kern-.025em b}\kern-.08em
    T\kern-.1667em\lower.7ex\hbox{E}\kern-.125emX}}
\usepackage{multirow}

\usepackage{amsmath, amsfonts, siunitx}
\usepackage{amssymb}
\usepackage{rotating}
\usepackage{array}
\usepackage{sidecap}

\usepackage{graphicx}
\usepackage[export]{adjustbox}

\usepackage{booktabs}

\usepackage{tikz}
\usetikzlibrary{shapes,arrows,backgrounds,fit,positioning}

\usetikzlibrary{decorations.markings}
\usetikzlibrary{arrows,automata}
\usetikzlibrary{positioning}
\usetikzlibrary{arrows.meta}
\usepackage{pgfplots}
\usepackage{color,soul}
\usepackage{caption}
\usepackage{stfloats}    
\usepackage{comment}
\usepackage{titlesec}
\titlespacing\section{0pt}{-1pt plus 0pt minus 0pt}{0pt plus 0pt minus 2pt}
\titlespacing\subsection{0pt}{-1pt plus 0pt minus 0pt}{0pt plus 0pt minus 2pt}
\titlespacing\subsubsection{0pt}{-1pt plus 0pt minus 0pt}{0pt plus 0pt minus 2pt}

\graphicspath{{img/}}
\usepackage{hyperref}
\usepackage{setspace}
\usepackage{pdflscape}
\usepackage{afterpage}
\usepackage{capt-of}
\title{Early Prediction of Vital Signs Using Generative Boosting via LSTM Networks}
\title{Generative Boosting: Early Prediction of Vital Signs via LSTM Networks}
\title{Long-range Prediction of Vital Signs Using Generative Boosting via LSTM Networks}

\author{%
  Shiyu Liu \normalfont{and} \textbf{Mehul Motani}\\
  Department of Electrical and Computer Engineering\\
  National University of Singapore\\
  \texttt{shiyu$\_$liu@u.nus.edu, motani@nus.edu.sg}  \\
}

\usepackage{fancyhdr}
\fancypagestyle{firstpage}{
  \fancyhf{}
  \fancyfoot[L]{\small Machine Learning for Health (ML4H) at NeurIPS 2019 - Extended Abstract}%
}

\begin{document}

\setlength{\abovedisplayskip}{0pt}
\setlength{\belowdisplayskip}{0pt}

\maketitle

\thispagestyle{firstpage}

\begin{abstract}
Vital signs including heart rate, respiratory rate, body temperature and blood pressure, are critical in the clinical decision making process. Effective early prediction of vital signs help to alert medical practitioner ahead of time and may prevent adverse health outcomes. In this paper, we suggest a new approach called generative boosting, in order to effectively perform early prediction of vital signs. Generative boosting consists of a generative model, to generate synthetic data for next few time steps, and several predictive models, to directly make long-range predictions based on observed and generated data. We explore generative boosting via long short-term memory (LSTM) for both the predictive and generative models, leading to a scheme called generative LSTM (GLSTM). Our experiments indicate that GLSTM outperforms a diverse range of strong benchmark models, with and without generative boosting. 
Finally, we use a mutual information based clustering algorithm to select a more representative dataset to train the generative model of GLSTM. This significantly improves the long-range predictive performance of high variation vital signs such as heart rate and systolic blood pressure. 
\end{abstract}

\section{Introduction}
Nowadays, well-developed artificial intelligence techniques provide opportunities to effectively utilize resourceful medical data and better inform clinical decision making \citep{Neill2013,Zhou2017BCB,Jia2017BIBM,shiyu,Schaar2017,Joseph2017,shiyuBIBM2019}. One practical application is early prediction of patients' vital signs as abnormal vital signs are often associated with certain health problems \citep{VitalSigns}. Effective early predictions help to alert medical practitioners ahead of time and may prevent adverse health outcomes \citep{Gang2015, Edward2017}. Such application makes machine learning in healthcare more impactful than other domains (e.g., economic) as the early prediction may save patients' lives through appropriate early intervention. For example, if we can detect abnormal heart rate of a heart disease patient 60 minutes in advance, it means this patient may have a cardiovascular event 60 minutes later. The clinician can ask the patient to take aspirin (75 - 150mg) now and aspirin will cause inhibition of platelet function occurring within 60 minutes of ingestion \citep{aspirin} and may result in highest plasma level of the drug being reached before the cardiovascular events, leading to decreased risk of adverse health outcomes. In addition to heart disease, such early prediction could help clinicians to do early treatment for other diseases as well. 

This paper contributes to early prediction of vital signs as follows:
\begin{enumerate}[label = $\bullet$, topsep=0pt, itemsep=-0ex, partopsep=0ex, parsep=0ex, itemindent=0.5cm, leftmargin=0cm]
	\item We propose a new approach called generative boosting which consists of a generative model to generate synthetic data for next few time steps, and several predictive models, to make long-range predictions based on observed and generated data. 		
	\item We explore generative boosting using long short-term memory (LSTM) for both the predictive and generative models, leading to a scheme called generative LSTM (GLSTM). Our experiments indicate that GLSTM outperforms a diverse range of strong benchmark models, with and without generative boosting. We find improvements of up to 30\% over the benchmark models.
	\item We propose a mutual information (MI) based unsupervised clustering algorithm to effectively select a group of representative patients, in order to better train the generative model of GLSTM. Our results demonstrate this approach gives an approximately 10\% improvement in long-range predictions of heart rate and systolic blood pressure over GLSTM.
	\end{enumerate}
	
\section{Generative Boosting via LSTM Networks}

\subsection{Direct \& Iterative Forecasting}
\label{sec2.1}
Approaches to long-range prediction (also called multistep-ahead forecasting) consists of two main classes: direct forecasting and iterative forecasting \citep{Hamz2009,Sorja2007,Bon2012}. In the direct forecasting, various independent models are trained to make predictions at different time steps. Assume we use past 3 observations $\{X_{t}, X_{t-1}, X_{t-2}\}$, we can write direct forecasting as \eqref{eq3} - \eqref{eq5}
\noindent\begin{minipage}{.5\linewidth}
\begin{align}
\hat{X}_{t+1} & = f_{d_1} (X_{t}, X_{t-1}, X_{t-2}, \theta_{d1}),   \label{eq3} \\
\hat{X}_{t+2} & = f_{d_2} (X_{t}, X_{t-1}, X_{t-2}, \theta_{d2}),   \label{eq4} \\
\hat{X}_{t+3} & = f_{d_3} (X_{t}, X_{t-1}, X_{t-2}, \theta_{d3}),   \label{eq5} \\
&\hdots \nonumber
\end{align}
\end{minipage}%
\begin{minipage}{.5\linewidth}
\begin{align}
\hat{X}_{t+1} & = f_{iter} (X_{t}, X_{t-1}, X_{t-2}, \theta_{iter}),   \label{eq6} \\
\hat{X}_{t+2} & = f_{iter} (\hat{X}_{t+1}, X_{t}, X_{t-1}, \theta_{iter}),   \label{eq7} \\
\hat{X}_{t+3} & = f_{iter} (\hat{X}_{t+2}, \hat{X}_{t+1}, X_t,\theta_{iter}),   \label{eq8} \\
&\hdots \nonumber
\end{align}
\end{minipage}
where $\hat{X}_{t+1}$, $\hat{X}_{t+2}$ and $\hat{X}_{t+3}$ denote the predictions for next three time steps using different models $f_{d_1}, f_{d_2}, f_{d_3}$ with parameters $\theta_{d1}, \theta_{d2}, \theta_{d3}$. 
However, the long-range predictive performance (e.g., $N$-step ahead) becomes worse as $N$ grows. In iterative forecasting, the same model $(f_{iter})$ is used iteratively and previous predictions are used together with the original data as new inputs to evaluate the next prediction. Using the same notations, we can write iterative forecasting as \eqref{eq6} - \eqref{eq8}. This approach effectively reduces the prediction horizon $N$, but it suffers from the error propagation which means small error in current prediction causes larger error in subsequent predictions \citep{taieb2012,Sorja2007}. 

\subsection{Generative Boosting \& Generative LSTM (GLSTM)}
\label{sec2.2}
In this subsection,  we propose a new approach called generative boosting, which consists of a generative model to generate synthetic data for next few time steps, and several predictive models, to make long-range predictions based on observed and generated data. Using the same notations  in Section \ref{sec2.1},  we can write generative boosting as
\noindent\begin{minipage}{.5\linewidth}
\begin{align}
\hat{X}_{t+1} & = f_{gen} (X_{t}, X_{t-1}, X_{t-2}, \theta_{gen}), \label{eq0}\\
\hat{X}_{t+2} & = f_{gen} (\hat{X}_{t+1}, X_{t}, X_{t-1}, \theta_{gen}), \label{eq00}\\
& \nonumber
\end{align}
\end{minipage}%
\begin{minipage}{.5\linewidth}
\begin{align}
\hat{X}_{t+3} & = f_{pred1} (\hat{X}_{t+2}, \hat{X}_{t+1}, X_{t},\theta_{pred1}),  \label{eq2} \\
\hat{X}_{t+4} & = f_{pred2} (\hat{X}_{t+2}, \hat{X}_{t+1}, X_{t},\theta_{pred2}),  \label{eq22} \\
&\hdots \nonumber
\end{align}
\end{minipage}

Generative boosting consists of a generative model $f_{gen}$, to generate synthetic data for the next time step via iterative forecasting (see \eqref{eq0} - \eqref{eq00}), and several predictive models $f_{pred}$, to make long-range predictions based on observed and generated data via direct forecasting (see \eqref{eq2} - \eqref{eq22}). Generative boosting attempts to strike a balance between direct forecasting and iterative forecasting. Thus, it may inherit advantages of both methods, leading to better long-range predictive performance. 
We replace $f_{gen}$ and $f_{pred}$ with LSTM, leading to a scheme called generative LSTM (GLSTM). This configuration is justified by the experimental results in \citep{shiyuBIBM2019}.

\section{Performance Evaluation}
\label{existing}

\subsection{Dataset Description}
\label{DataDes}
The dataset used in this paper contains information from 177 medical patients from a local hospital. The data collection for this study was approved by the local institutional ethics committee. The data includes both demographic information and medical information recorded by a wearable system which continuously monitors patients' vital signs. According to the completeness of data, 7 features are shortlisted and used in this paper: age, gender, heart rate, respiratory rate, blood oxygen saturation, body temperature and systolic blood pressure. 
The majority of vital signs are recorded at a regular interval of 5 minutes over a duration of 24 hours (on average). A small amount of missing values (less than 50 readings) are imputed using their last historical readings. Finally, we scale all variables to [0,1] and reshape all scaled data via a sliding observation window (size = number of time steps ($M$ = 20) $\times$ number of features ($K$ = 7)) into a three dimensional dataset $\mathbb{Q} \in \mathbb{R}^{S \times M \times K}$.

\begin{algorithm}[t!]
    \caption{\small Training \& Testing Procedures for GLSTM} 
     \label{algorithm1}
    \begin{algorithmic}[1]
    \REQUIRE \small Training Dataset $\mathbb{P}$ $(40\%)$, Training Dataset $\mathbb{G}$ $(20\%)$, Validation Dataset $(20\%)$, Test Dataset $(20\%)$; \vspace{-3mm}
    \STATE \small Training Dataset $\mathbb{G}$ $\xrightarrow[\text{}]{\text{train}}$ Generative Model, Validation Dataset $\xrightarrow[\text{}]{\text{tune}}$ Generative Model;
    \REPEAT
    \STATE Generative Model $\xrightarrow[\text{}]{\text{generate}}$ Synthetic Data of Next Time Step for Dataset $\mathbb{P}$ \& Test Dataset; 
    \STATE \small Remove Data of Oldest Time Step from Dataset $\mathbb{P}$ \& Test Data;
    \STATE \small Concatenate Generated Synthetic Data to Dataset $\mathbb{P}$ \& Test Data as Latest Time Step;
    \UNTIL {Generation Requirement is Met (e.g., up to next three time steps)}
    \STATE \small Modified Training Dataset $\mathbb{P}$ $\xrightarrow[\text{}]{\text{train}}$ Predictive Model, Validation Dataset $\xrightarrow[\text{}]{\text{tune}}$ Predictive Model; \vspace{-1mm}
    \STATE \small Modified Test Dataset $\xrightarrow[\text{}]{\text{test}}$ Predictive Model, Calculate MSE \& MAPE; \vspace{-2mm}
  \end{algorithmic}
\end{algorithm}

\subsection{Experiment Set Up}
\label{ep1}
Four benchmark models: Autoregressive Integrated Moving Average (ARIMA) model \citep{ARIMA}, Support Vector Regression (SVR) \citep{SVM}, Gaussian Process Regression (GPR) \citep{GPR} and LSTM \citep{LSTM} are shortlisted as they are reported to provide significant performance in time-series data prediction \citep{Sato2013,Helf1996,Imhoff1997,Sap2009,Muller1997a}. 
In the experiments, the dataset $\mathbb{Q} \in \mathbb{R}^{S \times M \times K}$ is randomly divided into 3 subsets at the patient level: training dataset $\mathbb{T} \in \mathbb{R}^{S_1 \times M \times K}$ (60\%), validation dataset $ \in \mathbb{R}^{S_2 \times M \times K}$ (20\%), and testing dataset $\in \mathbb{R}^{S_3 \times M \times K}$ (20\%). For benchmark models, LSTM is trained using training dataset $\mathbb{T}$ and tuned via validation dataset. The MLP (three hidden layers with size [10,5,3]) and SVR (kernel=$rbf$) are trained using flatted training dataset $\mathbb{T}$ ($\mathbb{R}^{S_1 \times M'} $) and tuned via flatted validation dataset ($\mathbb{R}^{S_2 \times M'} $) where $M' = M \times K$. The ARIMA (2,0,1) is trained via past 20 historical values of a single vital sign (e.g., HR). For GLSTM, the training dataset $\mathbb{T} \in \mathbb{R}^{S_1 \times M \times K}$ (60\%) are randomly divided into 2 subsets at the patient level: training dataset $\mathbb{P} \in \mathbb{R}^{S_4 \times M \times K}$ (40\%) and training dataset $\mathbb{G} \in \mathbb{R}^{S_5 \times M \times K}$ (20\%). The training and testing procedures are summarized in Algorithm \ref{algorithm1}. We note that both generative and predictive models contain one hidden layer with one LSTM unit and they are trained using batches of 20, sigmoid activation function and Adam \citep{adam} with a learning rate of 0.0005 and 0.001 for 300 epochs and 100 epochs, respectively. All parameters are tuned via validation dataset. Finally, we evaluate the performance of GLSTM and direct forecasting (via benchmark models) using testing dataset. All results, including GLSTM and benchmark models, are averaged over 10 runs. Two metrics, Mean Absolute Percentage Errors (MAPE) and Mean Squared Errors (MSE) are used to evaluate the predictive performance. In addition, Codes for reproducibility is available at \url{https://github.com/Martin1937/Generative-Boosting}.

\subsection{Performance Comparison}
\label{PC}
We now present experimental results of predicting heart rate (HR) at different time steps using proposed GLSTM (rows 5-7) and direct forecasting via benchmark algorithms (rows 1-4) in Table \ref{tab:hr}. The term following GLSTM, namely G1, G2 and G3, indicates the number of time steps for which synthetic data is generated. For example, in GLSTM-G1, the synthetic data is generated for the next time step (t+1) only and direct forecast predictions are made for subsequent time steps. In Table \ref{tab:hr}, the predictive performance for the time steps that are part of the generative model are not shown. For example, for GLSTM-G1, since the value at t+1 is part of the generative model, the predictive performance at t+1 is not shown.
For direct forecasting via benchmark models, we observe that LSTM outperforms all other benchmark models studied (rows 1-4 in Table \ref{tab:hr}). This finding agrees with the result stated in  \citep{Ma2015,Zachary2017,Gers2002,Malhotra2015}. Moreover, the proposed GLSTM has an improvement of at least 10\% when comparing to the best performing benchmark model (e.g., LSTM).

\begin{table}[t]
\small
\resizebox{1.0\textwidth}{!}{
    \setlength\tabcolsep{1pt} 
        \begin{tabular}{ c c c c | c c | c c | c c |c c| c c |c c}
	\cmidrule{1-14}
                &&\multicolumn{2}{c|}{t+1}&\multicolumn{2}{c|}{t+2}&\multicolumn{2}{c|}{t+3}&\multicolumn{2}{c|}{t+4}&\multicolumn{2}{c|}{t+5}&\multicolumn{2}{c|}{t+6}  \\ \cmidrule{1-14}
                &&  MSE&  MAPE & MSE& MAPE& MSE& MAPE& MSE& MAPE& MSE&  MAPE&  MSE& MAPE \\ \cmidrule{1-14}
                \multicolumn{1}{c}{\multirow{7}{*}}   &
                \multicolumn{1}{l}{{SVR}}& 65.26&8.78&69.68 & 8.99 & 73.64 & 9.23 & 76.85 &9.40&80.61&9.63&87.68&9.88 \\
                \multicolumn{1}{c}{}    &
                \multicolumn{1}{l}{GPR}& 62.88 & 7.45 & 68.26 & 8.08 & 72.03& 8.53 & 78.42 &9.29& 81.27&9.62&85.52&9.69 \\
                \multicolumn{1}{c}{}    &
                \multicolumn{1}{l}{ARIMA} &85.92 & 8.30 & 89.98 & 8.60 &92.94&8.85&95.03&8.99&97.71&9.14&99.83&9.26 \\
                \multicolumn{1}{c}{}    &   
                \multicolumn{1}{l}{LSTM} &{\textbf{\color{red}46.91}} & {\textbf{\color{red}6.69}} & 57.91 & 7.52 & 64.25&7.85&71.08&8.52&75.07&8.74&79.78&9.01 \\ \cmidrule{1-14}
                \multicolumn{1}{c}{}    &
                \multicolumn{1}{l}{GLSTM-G1} &-- &--&51.37&7.02&55.07&7.33&63.72&8.03&72.23&8.45&75.39&8.91 \\
                \multicolumn{1}{c}{}    &
                \multicolumn{1}{l}{GLSTM-G2} & --& -- & -- & --&58.37&7.58&63.95&7.91 &70.31&8.32&74.22&8.73 \\
                \multicolumn{1}{c}{}    &
                \multicolumn{1}{l}{GLSTM-G3} & -- & -- & -- & --&--&--&68.25&8.33 &74.56&8.72&79.42&9.11\\ \cmidrule{1-14}
                \multicolumn{1}{c}{}    &  
                \multicolumn{1}{l}{GLSTM-G1-MI} &-- &--&{\textbf{\color{red}47.31}}&{\textbf{\color{red}6.67}}&54.27&7.04&61.53&7.47 & {\textbf{\color{red}64.52}} & {\textbf{\color{red}7.81}} & 68.24 & 8.02 \\
                \multicolumn{1}{c}{}    &
                \multicolumn{1}{l}{GLSTM-G2-MI} & --& -- & -- & --&{\textbf{\color{red}53.78}}&{\textbf{\color{red}6.97}}&{\textbf{\color{red}60.23}}&{\textbf{\color{red}7.41}} & 65.25 & 7.84 & {\textbf{\color{red}67.22}} & {\textbf{\color{red}7.98}} \\
                \multicolumn{1}{c}{}    &              
                \multicolumn{1}{l}{GLSTM-G3-MI} & -- & -- & -- & --&--&--&65.74&7.96 & 69.73 & 8.35 & 73.51 & 8.62\\
	\cmidrule{1-14} 
        \end{tabular}    
        }
        \captionsetup{justification=raggedright,skip=0pt}
        \caption{Long-range prediction of Heart Rate (HR) with direct forecasting via various models (rows 1 -  4), GLSTM (rows 5 - 7), GLSTM with MI based clustering algorithm (rows 8 - 10). MSE and MAPE are the mean squared error and mean absolute percentage error, respectively. The best performance in each column is highlighted in red \& bold.}
        \label{tab:hr}
        \vspace{-7mm}
\end{table}

\section{Improving GLSTM via a Mutual Information based Clustering}
\label{section5_2}
Up to this point, we have assumed that training dataset $\mathbb{G}$ is randomly selected from training dataset $\mathbb{T}$, and hence may not represent dataset $\mathbb{T}$ well. For example, the dataset $\mathbb{T}$ contains patients from region $A$, $B$ and $C$. But randomly selected dataset $\mathbb{G}$ may only contain patients from region $A$. Hence, the dataset $\mathbb{G}$ may not be a good representative of dataset $\mathbb{T}$. Consequently, the generative model trained by dataset $\mathbb{G}$ may not learn the full distribution and results in poor generative performance. In this subsection, we propose a mutual information (MI) based unsupervised clustering algorithm to select a group of representative patients to better train the generative model. 

Mutual information $I (X;Y)$ from information theory \citep{cover2006elements} can quantify the dependency between random variable $X$ and $Y$. The higher the value, the better one variable can represent the other. The first step of our proposed MI based algorithm is assigning a score to each patient and the scoring function is given by $J(P_i) = \sum_{P_j \in \Omega, P_j \neq P_i }^{} I(P_i, P_j)$,
where $P_i \in \mathbb{R}^{p \times q}$ refers to the data of candidate patient with number of time steps $p$ and number of features $q$. We denote $\Omega$ as the dataset containing all patients (i.e., $\Omega = \{P_1, P_2, \cdots  \}$). The data length ($p$) of patients maybe different due to length of stay. Therefore, we estimate the MI via nearest-neighbor based MI estimators \citep{Kraskov2003}. After scoring each patient, we divide all patients into $L=10$ groups (of approximately equal size) based on the descending order of all scores. Therefore, patient data with similar scores will be grouped together and patient data within the same group will tend to be highly dependent on each other. Random sampling from each group will select representative samples of each group. Random proportional sampling from all groups gives a new training dataset $\mathbb{G'}$ and the rest of the patients form the new training dataset $\mathbb{P'}$. For fair comparison, we sample an equal amount as before (training dataset $\mathbb{G'}$: 20$\%$, training dataset $\mathbb{P'}$: 40$\%$). In such manner, we argue that we can select more representative training datasets for both the generative model and predictive models.

We now train the generative model (LSTM) and predictive model (LSTM) in GLSTM using dataset $\mathbb{G'}$ and dataset $\mathbb{P'}$, respectively. The performance of GLSTM with MI based clustering algorithm evaluated using the same testing dataset in Section \ref{PC} is shown in Table \ref{tab:hr} (row 8 - 10). The notation GLSTM with a suffix of MI means the model uses the MI clustering algorithm, otherwise the generative model and predictive model are trained using dataset $\mathbb{G}$ and dataset $\mathbb{P}$. In Table \ref{tab:hr}, we highlight the best performance in red \& bold. It can be seen that the predictive performance of GLSTM-MI has an approximately 10\% improvement as compare to GLSTM.

\section{Future Work}
\label{FutureWork}
We conclude the paper by presenting results for other vital signs and discussing clinical relevance.
\begin{itemize}[label = $\bullet$, topsep=0pt, itemsep=1ex, partopsep=0ex, parsep=0ex, itemindent=0cm, leftmargin=0.35cm]

\item {\bf Additional Experimental Results:}
To further evaluate the performance of GLSTM and the MI based clustering algorithm, we conduct similar experiments to perform long-range prediction of SBP (systolic blood pressure). The results shown in Appendix \ref{appendix} Table \ref{tab:bt_more} match those for prediction of HR. Experimental results of predicting HR for more time steps in the future are also shown in Appendix \ref{appendix}. Our best results indicate that our proposed model is able to predict HR and SBP 20 minutes in advance with a MAPE of 7.41\% and 6.17\%, respectively. For 60 minutes early prediction, the corresponding MAPE are 9.25\% and 7.33\%, respectively (see Table \ref{tab:hr_more} - Table \ref{tab:bt_more} in Appendix \ref{appendix}).  As expected, the MAPE increases with the prediction horizon. 

\item {\bf Feedback From Clinicians:}
We have discussed with clinicians about the relevance of our results. The general feedback is that early prediction (with a 30-60 minute lead time) is very relevant for ICU and high risk patients. Another point made was that the proposed method could be integrated into early warning systems (such as NEWS2 \citep{NEWS2}) to do early prediction of mortality or sepsis.
\end{itemize}

\section*{Acknowledgment}
This research was supported by the National Research Foundation Singapore under its AI Singapore Programme (Award Number: AISG-GC-2019-002) and by the Singapore Ministry of Education under grants WBS R-263-000-D35-114 and WBS R-263-000-D64-114.








\bibliographystyle{unsrt}
\bibliography{sample-bibliographyy}
\newpage
\appendix
\begin{landscape}
\section{Supplementary Results for Section \ref{FutureWork}}
\label{appendix}

\begin{table}[h]
\Large
\resizebox{1.6\textwidth}{!}{
    \setlength\tabcolsep{1pt} 
        \begin{tabular}{ c c c c | c c | c c | c c |c c| c c |c c |c c | c c | c c | c c | c c c c }
	\cmidrule{1-26}
	\cmidrule{1-26}
                & & \multicolumn{13}{c}{  MSE $\&$ MAPE (in percentage) for Different Prediction Horizons }\\ \cmidrule{1-26}
                &&\multicolumn{2}{c|}{t+1}&\multicolumn{2}{c|}{t+2}&\multicolumn{2}{c|}{t+3}&\multicolumn{2}{c|}{t+4}&\multicolumn{2}{c|}{t+5}&\multicolumn{2}{c|}{t+6}&\multicolumn{2}{c|}{t+7}&\multicolumn{2}{c|}{t+8}&\multicolumn{2}{c|}{t+9}&\multicolumn{2}{c|}{t+10}&\multicolumn{2}{c|}{t+11}&\multicolumn{2}{c}{t+12}  \\ \cmidrule{1-26}
                &&  MSE&  MAPE & MSE& MAPE& MSE& MAPE& MSE& MAPE& MSE&  MAPE&  MSE& MAPE &  MSE& MAPE &  MSE& MAPE &  MSE& MAPE &  MSE& MAPE &  MSE& MAPE &  MSE& MAPE \\ \cmidrule{1-26}
                \multicolumn{1}{c}{\multirow{7}{*}}   &
                \multicolumn{1}{l}{{SVR}}& 65.26&8.78&69.68 & 8.99 & 73.64 & 9.23 & 76.85 &9.40&80.61&9.63&87.68&9.88&97.25&10.03&98.12&10.04&104.31&10.21&107.52&10.48 & 112.4 & 10.72 & 119.5 & 10.91 \\
                \multicolumn{1}{c}{}    &
                \multicolumn{1}{l}{GPR}& 62.88 & 7.45 & 68.26 & 8.08 & 72.03& 8.53 & 78.42 &9.29& 81.27&9.62&85.52&9.69 &99.23&10.17&100.51&10.13&105.27&10.32&109.21&10.44 & 114.2 & 10.83 & 120.6 & 11.21 \\
                \multicolumn{1}{c}{}    &
                \multicolumn{1}{l}{ARIMA} &85.92 & 8.30 & 89.98 & 8.60 &92.94&8.85&95.03&8.99&97.71&9.14&99.83&9.26&101.6&9.41&109.9&9.57&116.8&10.71&118.4&10.81 & 124.2 & 11.02 & 129.3 & 11.51 \\
                \multicolumn{1}{c}{}    &   
                \multicolumn{1}{l}{LSTM} &{\textbf{\color{red}46.91}} & {\textbf{\color{red}6.69}} & 57.91 & 7.52 & 64.25&7.85&71.08&8.52&75.07&8.74&79.78&9.01 &82.95 & 9.17 & 84.03& 9.33 & 89.33 &9.60 & 92.99 & 9.89 & 95.71 & 10.02 & 99.21 & 10.24 \\ \cmidrule{1-26}
                \multicolumn{1}{c}{}    &
                \multicolumn{1}{l}{GLSTM-G1} &-- &--&51.37&7.02&55.07&7.33&63.72&8.03&72.23&8.45&75.39&8.91 &74.71 & 8.67 & 81.83 & 9.20 & 82.87 & 9.29 & 84.78 & 9.39 & 88.47 & 9.51 & 92.41 & 9.72 \\
                \multicolumn{1}{c}{}    &
                \multicolumn{1}{l}{GLSTM-G1-MI} &-- &--&{\textbf{\color{red}47.31}}&{\textbf{\color{red}6.67}}&54.27&7.04&61.53&7.47 & {\textbf{\color{red}64.52}} & {\textbf{\color{red}7.81}} & 68.24 & 8.02 & {\textbf{\color{red}70.51}} & {\textbf{\color{red}8.31}} & {\textbf{\color{red}73.25}} & {\textbf{\color{red}8.41}} & {\textbf{\color{red}75.31}} & {\textbf{\color{red}8.43}} & {\textbf{\color{red}77.98}} & {\textbf{\color{red}8.72}} & {\textbf{\color{red}81.23}} & {\textbf{\color{red}8.93}} & {\textbf{\color{red}84.25}} & {\textbf{\color{red}9.25}}\\
                \multicolumn{1}{c}{}    &
                \multicolumn{1}{l}{GLSTM-G2} & --& -- & -- & --&58.37&7.58&63.95&7.91 &70.31&8.32&74.22&8.73 &79.17 & 9.01 & 80.33 & 9.08 & 84.65 & 9.35 & 85.80 & 9.41 & 90.21 & 9.58 & 94.32 & 9.81 \\
                \multicolumn{1}{c}{}    &
                \multicolumn{1}{l}{GLSTM-G2-MI} & --& -- & -- & --&{\textbf{\color{red}53.78}}&{\textbf{\color{red}6.97}}&{\textbf{\color{red}60.23}}&{\textbf{\color{red}7.41}} & 65.25 & 7.84 & {\textbf{\color{red}67.22}} & {\textbf{\color{red}7.98}} & 71.24 & 8.35 & 74.27 & 8.47 & 77.25 & 8.68 & 80.21 & 9.07 & 83.22 & 9.21 & 87.61 & 9.32 \\
                \multicolumn{1}{c}{}    &
                \multicolumn{1}{l}{GLSTM-G3} & -- & -- & -- & --&--&--&68.25&8.33 &74.56&8.72&79.42&9.11& 80.31 & 9.09 & 82.09 & 9.19 & 84.44 & 9.41 & 88.41 & 9.48 & 92.45 & 9.62 & 97.31 & 9.92 \\
                \multicolumn{1}{c}{}    &                
                \multicolumn{1}{l}{GLSTM-G3-MI} & -- & -- & -- & --&--&--&65.74&7.96 & 69.73 & 8.35 & 73.51 & 8.62 & 73.25 & 8.42 & 77.28 & 8.82 & 79.41 & 8.91 & 81.23 & 9.11 & 85.52 & 9.42 & 89.71 & 9.52\\
	\cmidrule{1-26} 
        \end{tabular}    
        }
        \captionsetup{justification=raggedright,skip=0pt}
        \caption{Long-range early prediction of Heart Rate (HR) with direct forecasting via various models, GLSTM, GLSTM with MI based clustering algorithm for next 60 minutes. MSE and MAPE are the mean squared error and mean absolute percentage error, respectively. The best performance in each column is highlighted in red \& bold.}
        \label{tab:hr_more}
        \vspace{-20mm}
\end{table}

\begin{table}[b]
\Large
\resizebox{1.6\textwidth}{!}{
    \setlength\tabcolsep{1pt} 
        \begin{tabular}{ c c c c | c c | c c | c c |c c| c c |c c |c c | c c | c c | c c | c c c c }
	\cmidrule{1-26}
	\cmidrule{1-26}
                & & \multicolumn{13}{c}{  MSE $\&$ MAPE (in percentage) for Different Prediction Horizons }\\ \cmidrule{1-26}
                &&\multicolumn{2}{c|}{t+1}&\multicolumn{2}{c|}{t+2}&\multicolumn{2}{c|}{t+3}&\multicolumn{2}{c|}{t+4}&\multicolumn{2}{c|}{t+5}&\multicolumn{2}{c|}{t+6}&\multicolumn{2}{c|}{t+7}&\multicolumn{2}{c|}{t+8}&\multicolumn{2}{c|}{t+9}&\multicolumn{2}{c|}{t+10}&\multicolumn{2}{c|}{t+11}&\multicolumn{2}{c}{t+12}  \\ \cmidrule{1-26}
                &&  MSE&  MAPE & MSE& MAPE& MSE& MAPE& MSE& MAPE& MSE&  MAPE&  MSE& MAPE &  MSE& MAPE &  MSE& MAPE &  MSE& MAPE &  MSE& MAPE &  MSE& MAPE &  MSE& MAPE \\ \cmidrule{1-26}
                \multicolumn{1}{c}{\multirow{7}{*}}   &
                \multicolumn{1}{l}{SVR}& 97.19 & 6.20 & 115.8 & 6.64 & 125.3&6.94&133.1&7.12&138.8&7.28&145.2&7.45&151.7&7.64&158.2&7.80&162.9 &7.94&168.8& 8.09 & 173.2 & 8.17 & 179.2 & 8.32  \\
                \multicolumn{1}{c}{}    &
                \multicolumn{1}{l}{GPR}& 102.1 & 6.31 & 119.4 & 6.71 & 127.5&7.05&131.7&7.07&141.0&7.39&149.9&7.55&156.3&7.66 &161.4&7.87&167.7&8.12 &171.7&8.21 & 178.4 & 8.33 & 183.4 & 8.44\\
                \multicolumn{1}{c}{}    &
                \multicolumn{1}{l}{ARIMA} &150.9 & 6.44 & 157.4 & 6.70 &162.4&6.88&167.7&7.07&172.5&7.23&177.4&7.36&183.0&7.57 &187.4&7.65&191.1&7.72&196.2&7.84 & 201.5 & 8.22 & 209.3 & 8.41\\
                \multicolumn{1}{c}{}    &   
                \multicolumn{1}{l}{LSTM} &\textbf{{\color{red}82.77}} &\textbf{{\color{red}5.13}} & 106.2 & 6.02 & 116.6&6.27&123.2&6.43&132.9&6.76&138.2&6.85&145.4&7.09  &149.3&7.21&155.3&7.38&163.2&7.64 & 168.4 & 7.88 & 171.8 & 8.03\\ \cmidrule{1-26}
                \multicolumn{1}{c}{}    &
                \multicolumn{1}{l}{GLSTM-G1} &-- &--&98.77&5.80&112.2&6.15&122.5&6.31&129.6&6.71&134.2&6.81&143.1&7.05 &147.9&7.17&150.9&7.23&156.6&7.43 & 161.2 & 7.54 & 163.4 & 7.67 \\
                \multicolumn{1}{c}{}    &
                \multicolumn{1}{l}{GLSTM-G1-MI} &-- &--&\textbf{{\color{red}84.98}}&\textbf{{\color{red}5.41}}&\textbf{{\color{red}96.49}}&\textbf{{\color{red}5.93}}&\textbf{{\color{red}104.5}}&6.18&\textbf{{\color{red}109.6}}&\textbf{{\color{red}6.30}}&\textbf{{\color{red}113.6}}&\textbf{{\color{red}6.35}}&\textbf{{\color{red}120.3}}&\textbf{{\color{red}6.61}} &\textbf{{\color{red}127.1}}&\textbf{{\color{red}6.79}}&\textbf{{\color{red}132.8}}&\textbf{{\color{red}6.93}}&\textbf{{\color{red}141.9}}&\textbf{{\color{red}7.21}} & 149.2 & 7.31 & 154.3 & 7.37 \\
                \multicolumn{1}{c}{}    &
                \multicolumn{1}{l}{GLSTM-G2} & --& -- & -- & --&109.9&6.05&121.9&6.32 &126.8&6.60&136.7&6.72&141.8&7.03 &144.9&7.13&147.4 &7.20 &154.6&7.39&158.2 & 7.51 & 164.3 & 7.59 \\
                \multicolumn{1}{c}{}    &
                \multicolumn{1}{l}{GLSTM-G2-MI} & --& -- & -- & --&98.25&5.95&106.3&\textbf{{\color{red}6.17}} &112.3&6.35&121.1&6.68&130.0&6.91 &135.8&7.04&142.5 &7.33&144.1&7.26 & {\textbf{\color{red}147.4}} & {\textbf{\color{red}7.29}} & {\textbf{\color{red}151.7}} & {\textbf{\color{red}7.33}}  \\
                \multicolumn{1}{c}{}    &
                \multicolumn{1}{l}{GLSTM-G3} & -- & -- & -- & --&--&--&126.0&6.59 &130.8&6.82&134.3&6.89&144.8&7.25 &148.8&7.21&153.3&7.19&157.6&7.49 & 164.3 & 7.62 & 169.5 & 7.81\\
                \multicolumn{1}{c}{}    &                
                \multicolumn{1}{l}{GLSTM-G3-MI} & -- & -- & -- & --&--&--&118.8&6.36 &125.3&6.84&128.2&6.85&133.9&7.05 &138.8&7.16&141.2&7.17&149.3&7.27 & 157.4 & 7.54 & 161.3 & 7.71 \\
	\cmidrule{1-26} 
        \end{tabular}    
        }
        \captionsetup{justification=raggedright,skip=0pt}
        \caption{Long-range early prediction of Systolic Blood Pressure (SBP) with direct forecasting via various models, GLSTM, GLSTM with MI based clustering algorithm for next 60 minutes. MSE and MAPE are the mean squared error and mean absolute percentage error, respectively. The best performance in each column is highlighted in red \& bold.}
        \label{tab:bt_more}
        \vspace{0mm}
\end{table}
    \end{landscape}

\end{document}